\documentclass{article}

\usepackage[final]{neurips_2023}
\usepackage[utf8]{inputenc} % allow utf-8 input
\usepackage[T1]{fontenc}    % use 8-bit T1 fonts
\usepackage{hyperref}       % hyperlinks
\usepackage{url}            % simple URL typesetting
\usepackage{booktabs}       % professional-quality tables
\usepackage{amsfonts}       % blackboard math symbols
\usepackage{nicefrac}       % compact symbols for 1/2, etc.
\usepackage{microtype}      % microtypography
\usepackage{xcolor}         % colors
\usepackage{graphicx}
\usepackage{enumitem}
\usepackage{caption}
\newtheorem{myDef}{Definition}

\definecolor{darkred}{RGB}{229, 14, 14} 
\definecolor{forestgreen}{RGB}{12,132,198}
 \definecolor{darkslateblue}{RGB}{0,0,0}
\DeclareUnicodeCharacter{2248}{}

\usepackage{lipsum}

\newcommand\blfootnote[1]{
  \begingroup
    \renewcommand{\thefootnote}{} % 移除编号
    \footnotetext{#1}
    \endgroup
}

\title{Why Does ChatGPT Fall Short in Providing Truthful Answers?}

 \author{Shen Zheng$^{*}$ $\quad$ Jie Huang$^{*}$   $\quad$ Kevin Chen-Chuan Chang \\
 University of Illinois at Urbana-Champaign, USA \\
 \texttt{\{shenz2, jeffhj, kcchang\}@illinois.edu}
}

\begin{document}

\maketitle

\begin{abstract}
Recent advancements in large language models, such as ChatGPT, have demonstrated significant potential to impact various aspects of human life. However, ChatGPT still faces challenges in providing reliable and accurate answers to user questions. To better understand the model's particular weaknesses in providing truthful answers, we embark an in-depth exploration of open-domain question answering.
Specifically, we undertake a detailed examination of ChatGPT's failures, categorized into: \textit{comprehension}, \textit{factuality}, \textit{specificity}, and \textit{inference}. We further pinpoint \textit{factuality} as the most contributing failure and identify two critical abilities associated with factuality: \textit{knowledge memorization} and \textit{knowledge recall}. 
Through experiments focusing on factuality, we propose several potential enhancement strategies. Our findings suggest that augmenting the model with granular external knowledge and cues for knowledge recall can enhance the model's factuality in answering questions.
\blfootnote{$^*$Equal contribution.}
% \footnote{\texorpdfstring{$^*$Equal contribution.}}
\end{abstract}

\section{Introduction}
ChatGPT/GPT-4~\citep{openai2022chatgpt,openai2023gpt4} has gained substantial recognition for its practical applications, providing useful and informative responses to a wide range of queries.
Recent studies have conducted comprehensive technical evaluations of ChatGPT on numerous NLP tasks, demonstrating that ChatGPT outperforms other models across various tasks \citep{bubeck2023sparks,bang2023multitask,jiao2023chatgpt,qin2023chatgpt}.

However, in spite of the impressive capabilities exhibited by ChatGPT, researchers have highlighted some challenges of ChatGPT, such as its inability to perform reliable reasoning \citep{bang2023multitask}, translate low-resource languages effectively \citep{jiao2023chatgpt}, solve complex mathematical problems \citep{frieder2023mathematical}, and provide accurate information \citep{bang2023multitask}.
% While some researchers have discussed these issues \cite{bang2023multitask,bubeck2023sparks}, 
While these shortcomings are documented, the specific limitations of ChatGPT that contribute to these challenges are not entirely clear in the existing literature.
% However, it remains unclear in these studies which specific abilities of ChatGPT are limited and contribute to these problems.
Taking question answering as a representative example:
Is the model's failure due to its inability to reason or a lack of knowledge to answer the question?
Is the issue a result of insufficient knowledge, or does the model struggle to recall the internal knowledge with the question?
Is the difficulty in recalling knowledge the root cause, or does the model have trouble understanding the question's context or intent?

% \textit{What abilities of ChatGPT are still insufficient for answering questions faithfully?}
% Consequently, we pose the following question: 
% \textit{Why does ChatGPT fall short in answering questions faithfully?}
% Our goal is to explore the underlying reasons why ChatGPT cannot provide accurate and reliable answers to questions and identify specific ability deficiencies that cause ChatGPT to struggle for faithfual question answering.
In this study, we delve into an in-depth exploration of the weakness of ChatGPT in the context of complex open-domain question answering, as this task aligns closely with users' everyday search demands and requires extensive knowledge as well as robust understanding and reasoning capabilities.  Our goal is to identify common failure modes of ChatGPT in providing truthful answers, pinpoint the specific abilities in which ChatGPT is deficient that contribute to these failures, and consider potential strategies for improvement.
% Consequently, we pose the following question:
% why does ChatGPT fall short in providing truthful answers, and how can this be mitigated?.

% In this study, we delve into an in-depth exploration of the weakness of ChatGPT in the context of complex open-domain question answering, as this task aligns closely with users' everyday search demands and requires extensive knowledge as well as robust understanding and reasoning capabilities.

% Consequently, we pose the following question: 
% \textit{Why does ChatGPT fall short in providing faithful answers, and how can this be mitigated?} We consider an answer to be faithful in open domain question answering if it is factual based on real-world knowledge \cite{maynez2020faithfulness, peshterliev2021conversational}, relevant to the question, and specific at an appropriate level \cite{adiwardana2020towards,huang2022can}.
% Our goal is to explore the common failures of ChatGPT, to identify specific ability deficiencies that contribute to ChatGPT's challenges in question answering and examine potential approaches for enhancement.

 To this end, we first employ a thematic analysis to analyze instances of ChatGPT's failures and categorize them into four primary error types: \textit{comprehension error}, \textit{factuality error}, \textit{specificity error}, and \textit{inference error}. We then pinpoint the \textit{factuality} deficiency as the primary failure and identify \textit{knowledge memorization} and \textit{knowledge recall} as critical abilities for answering questions with \textit{factuality}.  Furthermore, we propose several potential strategies to help mitigate these deficiencies. Our results indicate that ChatGPT's \textit{factuality} can be enhanced by supplying granular external knowledge and cues for knowledge recall. Our findings provide practical insights for developing more reliable question answering systems.

\section{Related Work}

A substantial body of research has been conducted on examining various aspects of ChatGPT, including its general evaluation \citep{bang2023multitask,qin2023chatgpt,ko2023chatgpt}, understanding abilities \citep{zhong2023chatgpt}, mathematical abilities \citep{frieder2023mathematical}, bug fixing performance \citep{sobania2023analysis}, out-of-distribution (OOD) behaviors \citep{wang2023robustness}, translation behaviors \citep{jiao2023chatgpt}, and question answering performance \citep{guo2023close,tan2023evaluation}. Although ChatGPT showcased compelling performance, the research community has surfaced several issues concerning its reasoning \citep{borji2023categorical,bang2023multitask}, factual accuracy \citep{bang2023multitask}, solving complex mathematical problems \citep{frieder2023mathematical, borji2023categorical} and ethical implications \citep{zhuo2023exploring, borji2023categorical,ray2023chatgpt}. However, these studies predominantly concentrate on the categorization and identification of common problems, with limited in-depth investigation into the underlying deficiencies that contribute to the failures. In this work, we identify common failures in question answering scenarios, delve into the fundamental ability shortcomings that lead to these errors, and proposes potential strategies to mitigate these failures based on our experimental insights.

\section{Models and Datasets} 
We focus on complex open-domain question answering, using two widely-used benchmark datasets: HotpotQA~\citep{hotpotqa} and BoolQ~\citep{clark2019boolq}, both of which use Wikipedia as their knowledge source.
We selected 200 questions from HotpotQA for analyzing the errors made by ChatGPTs. For factuality evaluation, we sampled an additional 500 questions from HotpotQA and 1000 questions from BoolQ. We evaluated the performances of both GPT-3.5 and GPT-4 using these datasets. To generate responses from GPT-3.5 and GPT-4, we utilized the public OpenAI API.\footnote{\url{https://openai.com/blog/openai-api}. We use the model gpt-3.5-0301 and gpt-4-0314. }

\section{ChatGPT's Failures}

\subsection{Thematic Analysis}

We examined the model's responses to 200 HotpotQA samples using thematic analysis \citep{braun2012thematic}, a method for identifying patterns or ``themes'' within data. The process starts by extracting preliminary ``codes'' from data, which are later assembled into broader themes. For the purpose of ensuring a rigorous and comprehensive thematic analysis, a two-annotator approach was employed. We asked two independent annotators, both proficient in the subject and experienced in qualitative analysis to independently review the data set. Initially, they worked separately to identify the codes, analyze, and report patterns (themes) within the data. Upon completion of their independent analyses, the annotators convened in a collaborative session to compare, discuss, and reconcile any discrepancies in their identified themes. After the inter-annotator discussion, we grouped them several themes and validated these themes with an extra 40 questions. We confirm a complete representation of errors can be achieved through these themes.

To quantify the degree of consistency between the annotators, we employed Cohen's Kappa $\kappa$ \citep{kappa}. This statistical measure is used to evaluate the reliability of agreement between two raters beyond what is expected by chance. The formula for Cohen's Kappa is given by:

\[ \kappa = \frac{(p_o - p_e)}{(1 - p_e)} \]

In this formula, \( p_o \) represents the observed proportion of agreement, \( p_e \) represents the expected proportion of agreement.

In our study, the computed inter-coder agreement was 0.8394, which suggests a high degree of consensus between the annotators. We finalize the four identified themes as: problem comprehension and intent, factual correctness, specificity level, and reasoning, referred to as \textit{comprehension error}, \textit{factuality error}, \textit{specificity error}, and \textit{inference error}.

\begin{table*}[t]
    \centering
    \small 
    \caption{Examples of four types of errors (text in brackets represents the groundtruth, text in \textcolor{forestgreen}{blue} indicates the entities compared and their information, text in \textcolor{darkred}{red} indicates where the model (GPT-3.5) makes mistakes).}
    \begin{tabular}{p{1.8cm}|p{4cm}|p{6.7cm}}
        \hline
         Type & Question and GroundTruth & Responses  \\
        \hline
         Comprehension  &  What is a rare breed of dog that was derived as a variant of Rat Terrier, \textcolor{forestgreen}{Shiloh Shepherd dog} or \textcolor{forestgreen}{American Hairless Terrier}? \color{darkslateblue}\textit{(American Hairless Terrier)} & One rare breed of dog that was derived as a variant of \textcolor{darkred}{Rat Terrier}, Shiloh Shepherd dog or American Hairless Terrier is the Teddy Roosevelt Terrier.\\
        \hline
        Factuality  & Are the \textcolor{forestgreen}{New Orleans Outfall Canals} the same length as the \textcolor{forestgreen}{Augusta Canal}? \color{darkslateblue}\textit{(yes)} & \textcolor{darkred}{No}... the \textcolor{forestgreen}{Orleans Canal} is approximately \textcolor{darkred}{22 miles (35 kilometers)} long ... The \textcolor{forestgreen}{Augusta Canal} ... is approximately \textcolor{darkred}{7 miles (11.3 kilometers)} long \\
        \hline
        Specificity  & 1. What profession do \textcolor{forestgreen}{K≈çb≈ç Abe} and \textcolor{forestgreen}{Agatha Christie} share? \textcolor{darkslateblue}{\textit{(playwright)}} 
        
        2. What genre do \textcolor{forestgreen}{Superheaven} and \textcolor{forestgreen}{Oceansize} belong to? \textcolor{darkslateblue}{\textit{(rock)}} & 
        1. \textcolor{darkred}{Author}, ...
        
        2. Superheaven and Oceansize are \textcolor{darkred}{not of the same genre}.
        \textcolor{forestgreen}{Superheaven} ... play a style of \textcolor{forestgreen}{alternative rock} ...
        \textcolor{forestgreen}{Oceansize} ... was a British \textcolor{forestgreen}{progressive rock} band ... \\
        \hline
        Inference & 1. Which band has more members, \textcolor{forestgreen}{Muse} or \textcolor{forestgreen}{The Raconteurs}? \textcolor{darkslateblue}{\textit{(The Raconteurs)}}
        
        2. Which is currently more valuable, \textcolor{forestgreen}{Temagami-Lorrain Mine} or \textcolor{forestgreen}{Meadowbank Gold Mine}? \textcolor{darkslateblue}{\textit{(Meadowbank Gold Mine)}}
        & 1. \textcolor{darkred}{Muse has more members than The Raconteurs}. 
        \textcolor{forestgreen}{Muse} is a British rock band with \textcolor{forestgreen}{three members} ... The Raconteurs ... \textcolor{forestgreen}{Raconteurs has four members}. 
        
        2. I \textcolor{darkred}{cannot provide the current valuation} ...
        \textcolor{forestgreen}{Meadowbank Gold Mine}... was producing gold at a rate of \textcolor{forestgreen}{approximately 220,000 ounces per year}. 
        On the other hand, \textcolor{forestgreen}{Temagami-Lorrain Mine} is a historic iron mine located in Ontario, Canada that \textcolor{forestgreen}{has been inactive for many years}. 
        \\
        \hline
    \end{tabular}
    \vspace{-2mm}
    
    \label{tab:exp}
\end{table*}
% \vspace{-3mm}

\textit{Comprehension errors} refer to the failure in comprehending the question context and intention. In our experiment, the model demonstrates proficiency in comprehending the question, but it exhibits shortcomings when faced with questions containing grammar mistakes or ambiguity. For instance, the question shown in the Comprehension row of Table~\ref{tab:exp} presents a challenge to the model due to an incorrect interrogative pronoun ``what'', which should be ``which''. Consequently, the model fails to recognize that the question is seeking a selection between the two items marked in \textcolor{forestgreen}{blue} and instead misinterprets it as a choice among all last three items.

% \subsection{Factualness Error}
A \textit{factuality error} occurs when a model lacks the necessary supporting facts to produce an accurate answer \citep{petroni2019language,lee2023factuality}. This may be due to the model's lacking knowledge of a particular entity, attribute, or event. The example in the Factuality row of Table~\ref{tab:exp} shows a mistake when the model has the incorrect knowledge about the length of two canals. While this type of error is straightforward, it accounts for a majority of errors in the model.

\textit{Specificity errors} occur when the model fails to answer a problem at the appropriate level of specificity \citep{adiwardana2020towards,huang2022can}. This can manifest as the model providing an answer that is too general or too specific. For instance, in the specificity row of Table~\ref{tab:exp}, for the first question, the correct answer is playwright, but if the model answers with author, it is not specific enough. Similarly, in the second question, the ground truth is rock, but the model predicts \textcolor{forestgreen}{Superheaven} as alternative rock and \textcolor{forestgreen}{Oceansize} as progressive rock, and claims that they do not share the same genre. It is also classified as specificity error.

% \subsection{Inference Error}
An \textit{inference error} occurs when a model possesses the necessary knowledge to answer a question, but fails to reason with the facts effectively to arrive at the correct answer~\citep{huang2022reasoning,wei2023chainofthought}. For instance, for the first question in the inference row of Table~\ref{tab:exp}, the model may know that \textcolor{forestgreen}{Muse} has three members and \textcolor{forestgreen}{The Raconteurs} has four members, but still incorrectly claims that \textcolor{forestgreen}{Muse} has more members. Additionally, the model may fail to make predictions based on commonsense. In the second question, although holding the knowledge that the \textcolor{forestgreen}{Meadowbank Gold Mine} is still producing gold and the \textcolor{forestgreen}{Temagami-Lorrain Mine} has been inactive for years, the model still fails to deduce that the former one is currently more valuable due to its ongoing production.

% We analyzed the model's responses to 200 HotpotQA samples, studying the errors using thematic analysis \cite{braun2012thematic}, a method for identifying patterns or ``themes'' within data. This typically initiates with the extraction of preliminary ``codes'' from raw data, which are then aggregated into broader themes. We began by identifying errors in the responses as initial codes, followed by organizing these codes into themes. The details of initial codes and themes is shown in the appendix. \TODO{Comptete this} We confirmed the sufficiency and coverage of these themes with 40 extra questions as validation set, revealing that even the least common theme comprises 18.1$\%$ of errors and a complete representation of errors can be achieved through these four identified themes. Consequently, we identified four primary error categories: failure in comprehending the problem and intent, factual correctness, level of specificity, and effective reasoning. We designate these errors and introduce a novel classification of error categories for open-domain QA in the era of ChatGPT, as \textit{comprehension error}, \textit{factualness error}, \textit{specificity error}, and \textit{inference error}, respectively. More details in error types are shown in Appendix~\ref{sec:appendix}.

% \subsection{Results}
\subsection{Results} 

\label{sec:res}

\begin{table}[tp!]
\begin{center}
\caption{Number of failures under different settings.}
\scriptsize
\setlength\tabcolsep{1.9pt}
\scalebox{1.3}{
\begin{tabular}{lcccccc}
\hline
           & \textbf{\#Correct} & \textbf{\#Wrong} &  \textbf{Comprehension}  &  \textbf{Factuality} & \textbf{Specificity} & \textbf{Inference} \\ 
\hline
GPT-3.5             & 130        & 70       & 11       & 46        & 6            & 7          \\
GPT-3.5+evi & 186        & 14       & 1        & 6        & 2             & 5           \\
GPT-4               & 149        & 51       & 6        & 37       & 3             &    5       \\ 
\hline
\end{tabular}
}
% \vspace{-2mm}

\label{tab:manaul}
\end{center}

% \vspace{-2mm}
\end{table}

We counted the frequency of errors across the four categories. We used GPT-3.5 as the baseline model by feeding it plain questions. To study the effect of providing external evidence and to investigate GPT-4's improvement, we also explored providing questions with gold evidence (accurate piece of information that provides a definitive answer to a question) to the GPT-3.5 model (GPT-3.5+evi) and plain questions to the GPT-4 model. The results are summarized in Table~\ref{tab:manaul}. 

Based on our experiments, we made the following observations:
1) Nearly half of the failures are due to \textit{factuality error}, followed by \textit{inference error}, \textit{comprehension error} and \textit{specificity error}.
2) Providing evidence not only addresses \textit{factuality} but also significantly mitigates \textit{comprehension} and \textit{specificity errors}.
3) The GPT-4 model demonstrates some improvements compared to GPT-3.5, particularly in addressing \textit{comprehension} and \textit{specificity errors}. However, \textit{factuality} is only marginally improved. Our finding underlines \textit{factuality} as the primary concern in open-domain QA for its dominance among errors, its impact on other error types, and GPT-4's inadequate improvement in this area.
% , but \textit{factualness} and \textit{inference errors} are only marginally improved.

\section{Abilities Behind Factuality}
\begin{table*}[tp]
    \caption{Examples of questions and responses on factuality related abilities (The second column demonstrates the erroneous question and response, and third column illustrates our prompt to test the two abilities. Text in \textcolor{forestgreen}{blue} indicates the model's responses, and text in \textcolor{darkred}{red} is where the model make a mistake).}

    \centering
    \small 
    
    \begin{tabular}{p{1.6cm}|p{3.5cm}|p{7.5cm}}
        \hline
        Abilities & Erroneous Responses & Test Conversation \\
        \hline
         Memorization  &  --> Are the New Orleans Outfall Canals the same length as the Augusta Canal?:
         
% \textcolor{forestgreen}{
\textcolor{darkred}{--> No}\textcolor{forestgreen}{...the Orleans Canal is approximately} \textcolor{darkred}{22 miles (35 kilometers)} \textcolor{forestgreen}{...The Augusta Canal...is approximately} \textcolor{darkred}{7 miles (11.3 kilometers) long}

 & --> User (asking essential knowledge\#1): What's the length of New Orleans Outfall Canals?
 
 \textcolor{forestgreen}{--> Model: The total length of the New Orleans Outfall Canals is approximately}  \textcolor{darkred}{22.5 miles (36.2 kilometers).}

--> User (asking essential knowledge\#2): What's the length of Augusta Canal? 

\textcolor{forestgreen}{--> Model: The Augusta Canal is approximately} \textcolor{darkred}{7 miles (11.3 kilometers) long.}  \\
        \hline
        Recall  & --> Chang Ucchin was born in korea during a time that ended with the conclusion of what?
        
% \textcolor{forestgreen}{
\textcolor{darkred}{--> I'm sorry, but I need more information to answer your question accurately.} 
 & --> User (asking essential knowledge\#1): When was Chang Ucchin born in korea?

 \textcolor{forestgreen}{--> Model: Chang Ucchin ... was born on March 26, 1891...}

--> User (asking original question): Chang Ucchin was born in korea during a time that ended with the conclusion of what? 

\textcolor{forestgreen}{--> Model: Chang Ucchin was born in Korea during a time when Korea was under Japanese rule, ...ended with the conclusion of World War II in 1945.}

% \\    \hline
%         Reasoning  & --> Which band has more members, Muse or The Raconteurs?
        
% \textcolor{darkred}{--> Muse has more members than The Raconteurs.}
% \textcolor{forestgreen}{Muse is a British rock band with three members…
% The Raconteurs has four members.}
%  &  
        
% --> User (asking essential knowledge\#1): How many members do Muse have?

% \textcolor{forestgreen}{--> Model: Muse is an English rock band that was formed in 1994. The band currently consists of three members: ...}

% --> User (asking essential knowledge\#2): How many members do The Raconteurs have?

% \textcolor{forestgreen}{--> Model: The Raconteurs, also known as The Saboteurs in Australia, is an American rock band formed in 2005. The band consists of four members: ...}

% --> User (asking original question): Which band has more members, Muse or The Raconteurs?

% \textcolor{darkred}{--> Model: Muse has more members than The Raconteurs...}

\\

        \hline
    \end{tabular}
    % \vspace{-2mm}
    % \label{tab:exp}
    \label{tab:memo}
\vspace{-2mm}
\end{table*}
 
\label{sec:ability}

As we pinpoint in Section~\ref{sec:res} that \textit{factuality} is the most critical concern in open-domain QA, we focus on investigating factuality. Drawing from cognitive science research on human memory organization in QA \citep{lehnert2022process}, we treat \textit{factuality error} as a cognitive failure in retrieving relevant knowledge from LLMs' memory for a question.  

We define \textit{essential knowledge} $p$ with respect to a question $q$ as \textit{the knowledge indispensable for answering question $q$} and identify two key abilities for the knowledge retrieval process:

\begin{myDef}
(Knowledge Memorization) There exist an appropriate prompt $s$ which, when fed into the model, will result in the essential knowledge $p$.
\end{myDef}

\begin{myDef}{(Knowledge Recall)
Given the question $q$ as the prompt, the model is able to output the memorized essential knowledge $p$.}
\end{myDef}

Focusing on these two abilities, we conducted experiments with questions the model couldn't answer due to lack of knowledge. To test knowledge memorization, we rephrased essential knowledge as a question. E.g., in Table~\ref{tab:memo}, we evaluated the model's memorization by asking about canal lengths. To further evaluate knowledge recall, we re-asked the original question in the same conversation. 
% We concentrated on the above abilities and conducted experiments using questions that the model was unable to answer due to insufficient knowledge. To evaluate knowledge memorization, we reformulated essential knowledge into a question and presented it to the model. For example, in the memorization row of Table~\ref{tab:memo}, we inquire about the lengths of the two canals to assess the model's memorization. To further assess knowledge recall, we then asked the original question in the same conversation. 
If the model answered accurately in this setting but not when only asked the original question, it indicated a recall issue. In the recall row of Table~\ref{tab:memo}, the model correctly answers the query about Chang's birth event after a preceding related question, but fails without this context.

We find that 6 out of 46 errors stem from the recall issues, while the rest come from memorization. Our novel insight that distinguishes knowledge memorization and recall offers a new perspective for addressing knowledge-related problems.
% Our results revealed that $70.2\%$ of knowledge-related errors were due to the inability to memorize knowledge, while $14.9\%$ occurred during the knowledge recall process. our novel identification and quantification of knowledge recall could provide a new dimension for understanding and addressing knowledge-related errors.
% \jie{@Shen: Please complete this part}

% =====================

% \noindent \textbf{Definition 1. } \textit{(Knowledge Association): Given a problem $q$, a model $f$ possesses association ability if it can associate $q$ with the relevant knowledge $e$ required to answer $q$, based on the implicit knowledge contained in $f$.}

% \noindent \textbf{Definition 2. } \textit{(Reasoning): Given a problem $q$ and the associated or given knowledge $e$, a model $f$ possesses reasoning ability if it can use $e$ to predict the correct answer to $q$.}

% \subsection{Factualness}
\section{Improving Factuality for QA}

\subsection{Settings}

In our previous experiment, we demonstrated that a factuality error can stem from knowledge memorization or recall. To investigate how to mitigate it, we conducted experiments using the HotpotQA and BoolQ datasets. We used the plain question configuration with the GPT-3.5 model as our baseline. Since the GPT-3.5 and GPT-4 share similar underlying architectures and training processes, we experimented only on the former to draw observations. The prompts of our experiments are shown in Appendix~\ref{sec:prompt}. For evaluation, we employed partial match \citep{mavi2022survey}, which examines whether the ground truth is a substring of the prediction.

% In the following experiments, we provided external knowledge and context using the prompt: \texttt{Using the knowledge about [entity1, entity2, ...]. And with the following background knowledge [entity1: evidence1, entity2: evidence2, ...]. Answer the question: [question].} To encourage the model to generate concise responses, we added the prompt \texttt{Shorten the answer as much as possible} at the end. 

\subsubsection{Knowledge Memorization}

The limitations of models in memorizing facts has steered research towards retrieval-augmented language models, such as those augmented with external corpora \citep{izacard2022few,shi2023replug,huang2023raven} or search engines \citep{lazaridou2022internetaugmented,komeili2021internetaugmented}. However, noise and non-essential information can compromise retrieval, e.g., notable methods retrieve the whole web pages from Bing search as knowledge \citep{komeili2021internetaugmented}. Hence, we refer to the retrieved knowledge as context information and define \textit{granularity} as the length ratio of the context information to the essential knowledge. We investigate the role of \textit{granularity} in performance across four different granularity settings.

% When the model fails to memorize facts, providing external knowledge can be beneficial. Recent studies in retrieval-augmented prompting has become a popular research focus, including retrieving additional knowledge to combine with CoT prompting \cite{he2022rethinking, trivedi2022interleaving} and conditioning LMs on the retrieved information from Google Search \cite{lazaridou2022internetaugmented}. However, it is not always feasible to retrieve clean essential knowledge as sentences without any abundant information and noise. For example, \citet{lazaridou2022internetaugmented} retrieve the top 20 web page from google search and \citet{he2022rethinking} retrieves 10 relevant paragraphs from Wikipedia. Therefore, we are curious about how the \textit{granularity} in knowledge affects the performance, which, in our setting, is defined as: \textit{the proportion in length of provided knowledge that contains the key evidence}. We designed four different settings based on their granularity:
% \begin{itemize}
\begin{itemize}[leftmargin=*]
\item \textbf{Sentence level.} Directly provide external knowledge at the sentence level.

\item  \textbf{Passage level.} We offer gold evidence sentences along with other sentences related to the entities.

\item  \textbf{Section level.} We supply the Wikipedia section containing the gold evidence sentences.

\end{itemize}

\subsubsection{Knowledge Recall}

% In Section~\ref{sec:ability}, we noted that the model sometimes possesses knowledge but fails to recall it.

 To mitigate knowledge recall issues, we considered the knowledge recall process as retrieving the values (essential knowledge) in LLMs' memory with the given keys (the plain question by default), and tested whether supplying entity-related keys aids the knowledge recall process. Based on the keys provided, we proposed the following settings:

\begin{itemize}[leftmargin=*]
\item \textbf{Complete entity name.} We give the model complete Wikipedia names of the core entities in the question, e.g., for the film ``Samson and Deliah'', we provide ``Samson and Deliah (1984 film)''.

\item \textbf{Definition sentences.} We provide the entity's initial Wikipedia sentences as the definition or background in addition to the entity names, ensuring no essential knowledge is present.

\item \textbf{Random relevant sentences.} We also provide other random sentences from the entity's Wikipedia page along with entity names, again avoiding essential knowledge.
\end{itemize}

% In Section~\ref{sec:ability}, we observed that the model occasionally possesses knowledge but fails to recall that knowledge with the question. To address this problem, we investigated whether providing information relevant to the entity could help the model better recall the essential knowledge with the question. Based on different provided background, we designed the following settings:

% \begin{itemize}[leftmargin=*, nolistsep, nolistsep, topsep=1mm] \setlength{\itemsep}{1mm}
% \item \textbf{Complete Entity name.} Provide the model with the complete Wikipedia entity names of for the key entities in the question. For example, for the question about the film ``Samson and Deliah'', we supply the model with the complete entity name ``Samson and Deliah (1984 film)''.

% \item \textbf{Definition sentences.} In addition to the entity name, we also provided the first few sentences of the entity's Wikipedia page as the definition or background. To prevent essential evidence leakage, we verified that the evidence was not present in the selected background and replaced it if necessary.

% \item \textbf{Random relevant sentences.} As a contrast to the background setting, we selected several random sentences from the entity's Wikipedia page. We used the same approach to prevent key evidence leakage.

% \end{itemize}

\subsection{Observations and Implications}

% \begin{table}
% \centering
% \setlength\tabcolsep{5pt}
\begin{table}[tp!]
\begin{center}
\caption{Factuality experiments on HotpotQA and BoolQ.}

% \scriptsize
% \setlength\tabcolsep{4pt}
% \scalebox{1.3}{
\begin{tabular}{lcc}
\hline
                                      & HotpotQA                    & BoolQ \\ \hline
Plain Question                        & 0.58 & 0.71  \\
\hline
External - Sentence            & 0.73                        & 0.869 \\
External - Passage & 0.718                        & -     \\
External - Section   & 0.603                       & 0.77  \\
\hline
Recall - Complete Entity Name             & 0.597                       & 0.755 \\
Recall - Wiki Background         & 0.646                        & 0.789 \\
Recall - Random Sentences        & 0.55                       & -     \\
\hline
\end{tabular}
% }
\label{tab:fact}
% \end{table}
\end{center}

% \vspace{-10mm}
\end{table}

% \subsubsection{}
\textbf{\textit{Finding I. Finer granularity of external knowledge yields better results}}.
Table~\ref{tab:fact} (top) shows that external knowledge incorporation boosts performance, and effectiveness is influenced by knowledge granularity. Including essential knowledge with other sentences affects (although only slightly) results, while using the whole Wikipedia section with evidence greatly reduces the performance gain. This suggests that performance decreases with coarser input knowledge granularity.
% Table~\ref{tab:fact} presents our results regarding knowledge memorization and recall. Our findings indicate that incorporating evidence effectively improves the performance, and the granularity of the input knowledge significantly affects this effectiveness. We observed that combining key evidence with other sentences does not greatly affect the results, whereas providing the entire Wikipedia section containing the evidence tends to decrease accuracy. Performance declines as the granularity of the input knowledge decreases: the model equipped with the correct ground truth achieves the best performance, while providing the entire Wikipedia section yields a similar effect to simply providing the entity name.

\noindent \textbf{\textit{Finding II. Providing relevant keys aids in recalling essential knowledge}}.
% \subsubsection{Providing relevant keys aids in recalling essential knowledge}
Table~\ref{tab:fact} (bottom) indicates that supplying the model with keys can enhance accuracy. Supplying the complete entity names improves the performance, while providing entity background or definition sentences further aids knowledge recall, even without the essential knowledge. However, random sentences from the entities' Wikipedia page do not improve the  performance, but instead decrease it.
% Results in Table~\ref{tab:fact} (bottom) suggest that, even without external knowledge, supplying the model with the complete name of the key Wikipedia entity can improve accuracy to some extent. In addition to the entity name, providing background or definition sentences for the entity further assists the model in recalling with the key knowledge about the entity, even when the background sentences do not contain the key answer. However, we also discovered that randomly including sentences from the entity's Wikipedia page did not improve performance and made the model perform worse than merely providing the entity name.

% \subsection{Discussion}
% In this section, 
Based on these findings, we explore strategies to boost the factuality in question answering  from a LLM research perspective. 
\begin{itemize}[leftmargin=*]
    \item \textbf{Provide external context with fine granularity as evidence to help memorize essential knowledge.} Although LLMs possess a vast amount of implicit knowledge, it is evident that there is still a significant amount of knowledge that is difficult to cover during training or challenging to recall during inference. Therefore, building an IR system to retrieve knowledge with finer granularity based on the question would be helpful according to our findings. Some attempts in this direction have been observed in systems such as New Bing\footnote{\url{https://www.bing.com/new}}, Bard\footnote{\url{https://bard.google.com}}, and ChatGPT plugins\footnote{\url{https://openai.com/blog/chatgpt-plugins}}.
    \item \textbf{Provide descriptions of entities as keys to help recall essential knowledge.} 
    % While previous work has investigated the association abilities in pretrained language models \cite{huang-etal-2022-large}, there has been limited research on how to enhance these abilities in LMs. 
    In our analysis, we observe that sometimes ChatGPT indeed memorize the essential knowledge to solve the question, but cannot recall the knowledge with the question.
    A relevant finding is highlighted in the study by \citet{huang-etal-2022-large}, where it is observed that while language models can memorize a substantial number of email addresses, they struggle to associate specific email addresses with corresponding individual names.
    Improving the recall capabilities could be an intriguing direction for developing more powerful language models.
\end{itemize}

% For users, we recommend the following strategies:

% \begin{itemize}[leftmargin=*, nolistsep, nolistsep, topsep=1mm] \setlength{\itemsep}{1mm}
%     \item \textbf{Provide background information as possible.} Our research suggests that external background information can help the model recall more knowledge with the entity, thereby improving the ability to retrieve key facts and provide truthful answers. If no background information is available, simply providing the complete name of the entity as a hint can still have an effect.
%     \item \textbf{Provide external knowledge that is as specific as possible.} The granularity of external knowledge has been shown to affect the ability to retrieve facts and answer questions. Therefore, we recommend the users can provide external knowledge that is as precise as possible.
% \end{itemize}

\section{Conclusion}
Our study explores ChatGPT's common errors in the context of truthful open-domain question answering, identifies four error types and pinpoints \textit{factuality error} as the most critical error. We further define \textit{essential knowledge}, and examine two crucial abilities \textit{knowledge memorization} and \textit{knowledge recall} associated with factuality. We study the impacts of evidence granularity on \textit{knowledge memorization} and providing relevant keys on \textit{knowledge recall}. We finally propose methods to improve ChatGPT's factuality in QA, contributing to the understanding of factuality and offering insights to enhance QA systems and language models, promoting more reliable LLMs.

\clearpage
% \section*{Limitations}
% While this research brings forth some critical insights, it presents certain limitations.

% Firstly, our investigation into failures and factuality has been predominantly concentrated on two Wikipedia-based datasets: HotpotQA and BoolQ. It is important to note that despite our experimental focus on Wikipedia-sourced knowledge, the methodologies we propose can be applied to broader scenarios.  For example, knowledge memorization can be enhanced by external knowledge retrieved by a search engine, and knowledge recall can benefit from exploring online resources for definitions or complete names of entities. Despite the limitation, we consider this work as the initial attempt to critically characterize the factuality of ChatGPT, which might stimulate future work to delve into factuality in a wider range of scenarios.

% Secondly, our study focuses on factuality since it is demonstrated to be the most critical error. Nonetheless, other categories of failures such as \textit{inference error} are also worthy of scholarly attention. In particular, \textit{inference error} seems difficult to be improved by employing GPT-4 and providing external knowledge, as shown in Table~\ref{tab:manaul}. Some research has already focused on problem decomposition \cite{zhou2023leasttomost,khot2023decomposed,dua2022successive} to address this difficulty, which we believe is necessary and worthy.

\bibliography{custom}
\bibliographystyle{abbrvnat}

\clearpage
\appendix

\section{Prompts}
\label{sec:prompt}

\subsection{Knowledge Memorization}
The prompts for all the settings in knowledge memorization are:  

\texttt{Use the following knowledge [entity1: evidence1, entity2: evidence2, ...], answer the question: [question].}

\subsection{Knowledge Recall}
The prompt for complete entity name setting is:

\texttt{Use the knowledge about [entity1, entity2, ...], answer the question: [question].}

The prompts for the other two settings (definition sentences and random relevant sentences) are: 

\texttt{Use the knowledge about [entity1, entity2, ...], and with the following background knowledge [entity1: evidence1, entity2: evidence2, ...], answer the question: [question].}

\end{document}